\documentclass[letterpaper]{article} 
\usepackage{aaai2027} 
\usepackage[hyphens]{url} 
\usepackage{graphicx} 
\usepackage{natbib} 
\usepackage{caption} 
\usepackage{subcaption}
\usepackage{booktabs}
\usepackage{amsmath}
\usepackage{amssymb}
\usepackage[capitalize,noabbrev]{cleveref}
\usepackage{tcolorbox}

\title{HiWE: Hierarchical World Knowledge Model with Visual Keypoint Enhancement for Zero-Shot 3D Path Planning}
\author{
    Guoqing Ma $^{1,2,4}$, Mingqi Yuan $^{3}$, Chen Gao $^{5}$, Jiayu Chen $^{3}$, Shan Yu $^{1,4}$ \dag
  \\
\thanks{\dag Corresponding author: shan.yu@nlpr.ia.ac.cn} 
  $^{1}$ Institute of Automation, Chinese Academy of Sciences, Beijing, China\\
  $^{2}$ School of Future Technology, University of Chinese Academy of Sciences\\
  $^{3}$ The University of Hong Kong, Hong Kong SAR, China\\
  $^{4}$ State Key Laboratory of Brain Cognition and Brain-inspired Intelligence Technology, CAS\\
  $^{5}$ Department of Electronic Engineering, Tsinghua University,
China \\
}

\affiliations{}

\begin{document}

\maketitle

\begin{abstract}
Robot demonstration generation requires a system to identify where an interaction should occur, plan a feasible motion, and execute the required contact. HiWE connects these decisions through a point-based interface between visual grounding and language-based planning. PointVLM is instruction-tuned to associate task-relevant objects with image coordinates using a mixture of point annotations, segmentation-derived samples, robot observations, and visual question answering data. Depth measurements lift these predictions into a semantic 3D representation. A language planner, 3DLLM, uses this representation to specify end-effector waypoints and gripper commands, while a hybrid grasping module resolves local grasp poses. The evaluation covers 14 simulated manipulation tasks and four physical-robot tasks, together with ablations of the visual training data, spatial inputs, and grasp selection. Here, zero-shot execution refers to deployment without task-specific demonstration training; the visual model uses existing robot data during fine-tuning. This paper describes the original point-based formulation of the framework; its relationship to the subsequent GeneralVLA extension is detailed in the introduction.
\end{abstract}

\section{Introduction}
\label{sec:intro}
Collecting a successful robot demonstration requires more than recognizing an object or describing an intended action. A system must locate an interaction region, relate that region to the geometry of the scene, and execute a motion that respects the object and its surroundings. Errors at any of these stages can invalidate the resulting demonstration. This paper studies how an explicit point representation can connect visual grounding to motion planning for this purpose.

Action-predicting vision-language models, such as OpenVLA~\cite{DBLP:conf/corl/KimPKXB0RFSVKBT24}, learn from observations paired with robot actions. Large datasets expand the range of available training experience~\cite{DBLP:conf/icra/ONeillRMGPLPGMJ24,DBLP:conf/rss/KhazatskyP0BDKN24}, but collecting deployment-specific demonstrations remains costly. An alternative is to divide the problem into components with different supervision requirements: image-space grounding can use point or region annotations, a language model can reason over a spatial description, and a controller can handle execution. The usefulness of this division depends on what information passes between the components.

\begin{figure*}[t]
  \centering
  \includegraphics[width=.94\textwidth]{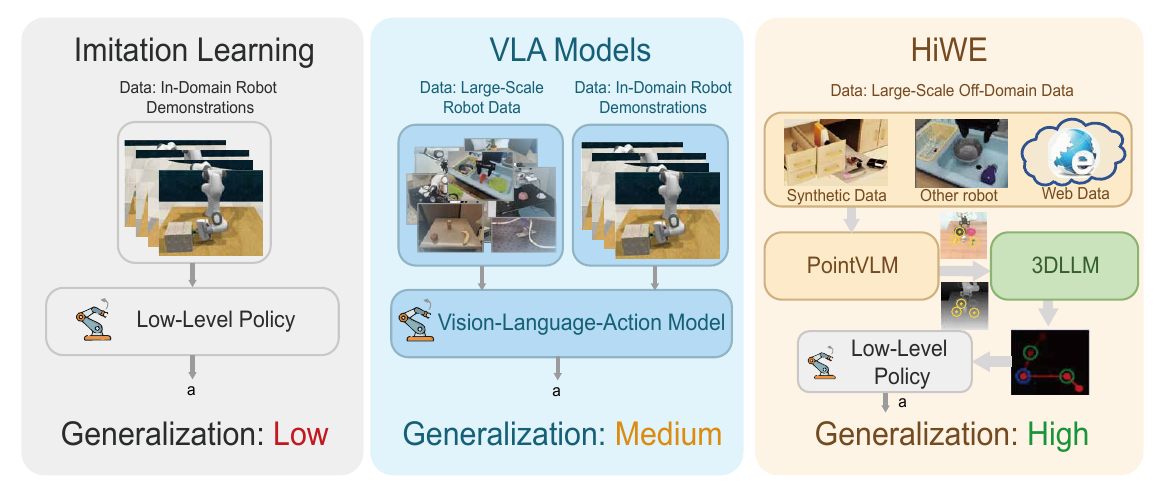}
  \caption{\textbf{Interfaces for robot manipulation.} HiWE exposes task-relevant points and planned trajectories between perception and execution. This separation allows the visual component to be trained on annotations that do not require deployment-specific action trajectories.}
  \label{fig:intro}
\end{figure*}

HiWE (\textbf{Hi}erarchical \textbf{W}orld Knowledge Model with Visual Keypoint \textbf{E}nhancement) uses PointVLM to predict object-associated image coordinates. Depth measurements convert these coordinates into 3D points for 3DLLM, which generates a sequence of waypoints and gripper commands. A hybrid grasping module (HGM) selects local grasp poses for execution. We use ``world knowledge'' to describe the pretrained models' semantic and spatial priors; this terminology does not imply that HiWE learns an explicit transition model of the environment.

\paragraph{Relationship to GeneralVLA.}
HiWE is the original formulation of the framework, and GeneralVLA~\cite{ma2026generalvla} is a subsequent extension that changes perception and adds experience-based planning. This relationship concerns the development of the methods, rather than their order of appearance on arXiv. HiWE uses PointVLM for direct coordinate prediction and 3DLLM for planning from the current instruction and scene. GeneralVLA introduces ASM, which combines language-conditioned segmentation with iterative refinement, and a KnowledgeBank that retrieves and accumulates experience across executions. The three-stage organization, the spatial planning interface, and HGM are common to the two manuscripts. We describe these common components here as part of the original system, while identifying segmentation refinement and persistent experience retrieval as features of the extension. Some experimental values and implementation descriptions also occur in both manuscripts, as noted alongside the relevant results.

\begin{table}[t]
\centering
\caption{Scope of the original system and its subsequent extension. Shared components are not independent evidence of a new architecture.}
\label{table:relationship}
\small
\begin{tabular}{p{0.20\columnwidth}p{0.34\columnwidth}p{0.34\columnwidth}}
\toprule
Component & HiWE & GeneralVLA \\
\midrule
Perception & PointVLM coordinate prediction & ASM segmentation and refinement \\
Planning & Current task and 3D scene points & 3D scene planning with KnowledgeBank \\
Experience & No persistent retrieval in 3DLLM & Retrieval, construction, consolidation \\
Execution & HGM grasp selection & HGM grasp selection \\
\bottomrule
\end{tabular}
\end{table}

The present evaluation examines point localization, the composition of PointVLM's training data, and the spatial information supplied to the planner. It also measures task execution in simulation and on a physical robot. Throughout this paper, ``zero-shot'' concerns deployment without task-specific demonstration training. It does not mean that every component is untrained or that no robot data are used: PointVLM is fine-tuned using existing datasets, including robot observations from outside the deployment environment.

This paper presents the original system and its evaluation through three contributions:
\begin{itemize}
  \item A point-based interface that connects an instruction-tuned visual model to a language planner and a grasp execution module, separating visual supervision from deployment-specific action demonstrations.
  \item A concrete implementation comprising PointVLM training, depth-based spatial grounding, 3DLLM waypoint generation, and HGM grasp selection.
  \item An evaluation on simulated and physical manipulation tasks, with ablations of visual training data, planner inputs, and grasp selection, and a study of demonstration-based policy training.
\end{itemize}
The experiments characterize this original configuration. They do not isolate the incremental benefit of GeneralVLA's segmentation or memory extensions; that would require a controlled comparison under matched conditions.

\begin{figure*}[t]
  \centering
  \includegraphics[width=\textwidth]{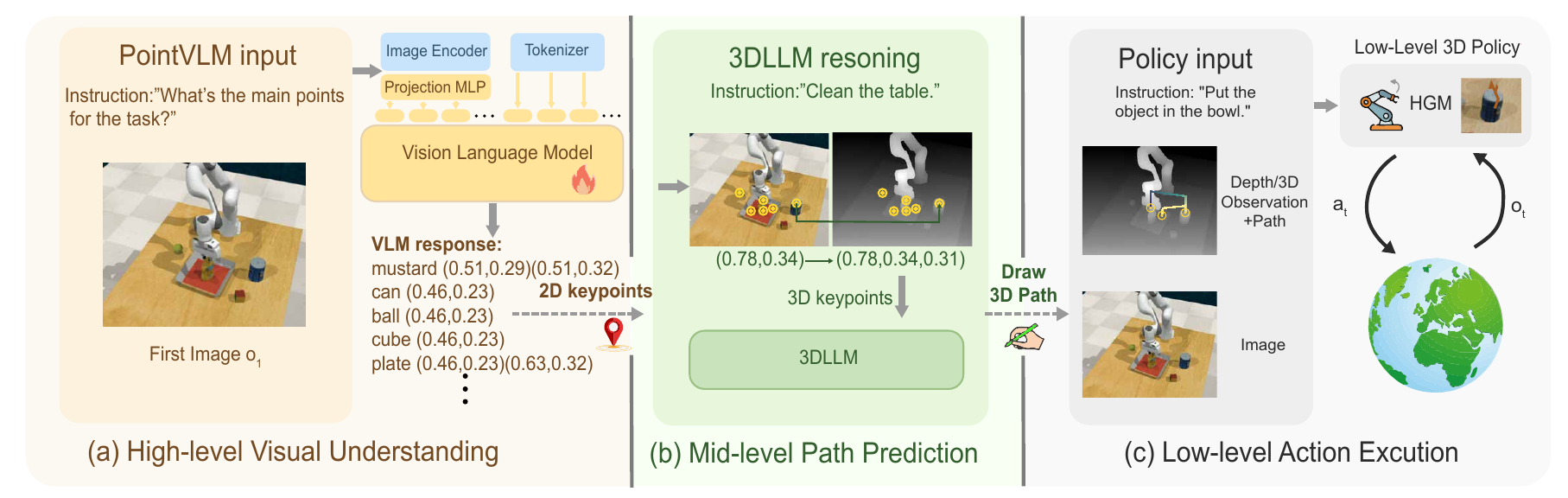}
  \caption{\textbf{HiWE data interfaces.} PointVLM associates task-relevant objects with image points. Depth supplies their 3D coordinates for 3DLLM, which outputs waypoints and gripper commands. HGM provides local grasp poses for the execution stage. The overall decomposition and grasping component are also described in GeneralVLA~\cite{ma2026generalvla}.}
  \label{fig:pipline}
\end{figure*}

\section{Related Work}
\noindent\textbf{Language models as robot planners.}
Code-as-Policies~\cite{DBLP:conf/icra/LiangHXXHIFZ23} represents a plan as executable code calling control functions. VoxPoser~\cite{DBLP:conf/corl/HuangWZL0023} connects language to geometric planning through 3D value maps, while Scaling-up~\cite{DBLP:conf/corl/HaFS23} uses language-guided interaction to obtain data for policy learning. These systems provide different interfaces between semantic decisions and robot execution. HiWE uses an ordered sequence of spatial waypoints and gripper commands; its evaluation compares the resulting system against these baselines under the inputs specified in the experiments.

\noindent\textbf{Learning actions from vision and language.}
Models including RT-1 and OpenVLA learn action predictions conditioned on observations and instructions~\cite{DBLP:conf/rss/BrohanBCCDFGHHH23,DBLP:conf/corl/KimPKXB0RFSVKBT24}. LLARVA additionally predicts trajectories as part of its learning formulation~\cite{DBLP:conf/corl/NiuSBQBSDH24}. HiWE instead exposes an intermediate spatial description to a separate planner and execution module. This changes the interfaces and supervision used by the system; it does not guarantee that reasoning, visual accuracy, or execution speed will improve in every setting.

\noindent\textbf{Points and trajectories as intermediate representations.}
Affordance prediction connects object semantics to possible interaction locations~\cite{DBLP:conf/corl/SundaresanBSB23,DBLP:conf/icml/NasirianyX0XL0X24,DBLP:conf/corl/YuanDBPKMMF24}. RoboPoint is particularly relevant to PointVLM because it supplies point-based visual supervision. Trajectory specifications also provide a way to condition policies~\cite{DBLP:conf/iclr/GuKW0AR0FGXSX0H24}, and point tracking makes motion information available from image sequences~\cite{DBLP:conf/iccv/DoerschYVG0ACZ23,DBLP:conf/corl/YuanWZG24,DBLP:conf/rss/WenLS0D0A24}. GeneralVLA~\cite{ma2026generalvla} is the closest architectural comparison because it combines semantic affordances, 3D planning, and grasp-pose selection. As the subsequent extension of HiWE, it retains the planning and execution backbone while developing the perception and experience-reuse mechanisms. The relationship is summarized in~\Cref{table:relationship}.

\section{HiWE: Point-Based Grounding and Planning}
\subsection{System interfaces}
HiWE passes explicit spatial information between perception, planning, and execution (\Cref{fig:pipline}). PointVLM receives an image and task instruction and returns object-associated image points. Depth supplies a 3D location for each point. The resulting semantic scene description is the input to 3DLLM, whose output consists of end-effector waypoints and gripper commands. HGM resolves local grasp poses before a motion planner executes the commands. GeneralVLA retains this overall organization while extending the perception and planning components~\cite{ma2026generalvla}.

\subsection{PointVLM: predicting interaction locations}
\label{HiWE’S PointVLM}
PointVLM provides the geometric anchors used by the subsequent planner. Given a task instruction, it identifies relevant objects and represents each object by a set of image coordinates, $\{object:(x_0,y_0),\ldots,(x_n,y_n)\}$. Associating several points with an object gives the planner information about its spatial extent after depth projection. The output is a coordinate sequence; HiWE does not use the segmentation decoder or segmentation-feedback loop described for ASM in GeneralVLA.

\noindent\textbf{Training.}
The visual backbone is QWen-VL-7B~\cite{DBLP:journals/corr/abs-2409-12191}. Following the instruction-tuning formulation of~\cite{DBLP:conf/nips/LiuLWL23a}, the manuscript's implementation updates the MLP projector and vision-language transformer while keeping the image encoder and tokenizer fixed. Responses are generated autoregressively, with a boundary token delimiting the instruction and answer. Point supervision specifies where an interaction can occur, while language supervision supports interpretation of the instruction.

\noindent\textbf{Supervision sources.}
The training mixture contains five sources (\Cref{fig:dataset}). RoboPoint~\cite{DBLP:conf/corl/YuanDBPKMMF24} contributes 347k point-prediction examples. For LVIS~\cite{DBLP:conf/cvpr/GuptaDG19}, points are sampled inside segmentation masks and paired with object semantics. Robot observations come from Open X-Embodiment~\cite{DBLP:conf/icra/ONeillRMGPLPGMJ24} and SIMPLER~\cite{DBLP:conf/corl/LiHGMPWFLSKL0F024}; these sources are outside the deployment environment. Finally, 667K visual question-answering conversations~\cite{DBLP:conf/iccv/KafleK17} provide language-based supervision. The contribution of each source is evaluated in~\Cref{table:datacomposition}. Additional implementation details appear in Appendix~\ref{Implementation and Architecture Details}.

\subsection{3DLLM: planning from a spatial description}
\label{Mid Level LLM}
Depth and camera geometry lift PointVLM's image coordinates into 3D. Each coordinate remains associated with its object name, so 3DLLM receives a task instruction together with a compact, semantic description of the scene. The planner specifies an ordered sequence of spatial waypoints interleaved with commands to open or close the gripper. This interface can express several interaction stages within one plan.

Multiple points help describe an object's orientation and extent; obstacle points supply context for choosing a route. These inputs are evaluated separately in~\Cref{table:3DLLM}. They inform the proposed motion but do not guarantee collision-free execution. Unlike the KnowledgeBank-equipped planner in GeneralVLA~\cite{ma2026generalvla}, the HiWE planner described here does not retrieve or consolidate a persistent collection of previous experiences. It plans from the current instruction and scene description. Training a separate behavior-cloning policy on successful demonstrations, described in the appendix, is distinct from such retrieval during planning.

\subsection{HGM: resolving local grasp geometry}
\label{Path Guided Low-level policy learning}
A waypoint specifies where the end effector should move, but a successful grasp also requires an appropriate orientation and contact configuration. HGM uses the task-relevant 3D points to restrict the region of interest in the RGB-D reconstruction. The grasp predictor~\cite{DBLP:conf/corl/YuanMMF23} then generates candidate poses in the resulting object-centered point cloud.

HGM rejects candidates that fail its collision check and selects the remaining candidate whose grasp center is nearest to the object center. This is a selection heuristic rather than a guarantee of a globally optimal grasp. The selected pose is combined with the planned waypoints and passed to the motion planner for execution. HGM is part of the original HiWE pipeline and also appears in the GeneralVLA extension; the ablation in~\Cref{table:HGM} evaluates its role within HiWE.

\begin{table*}[t]
\setlength{\tabcolsep}{4pt}
\caption{\textbf{Task-averaged success rate \% for zero-shot evaluation.} HiWE outperformed other baselines in 10 out of 14 simulation tasks from RLBench~\cite{DBLP:journals/ral/JamesMAD20}. Each task was evaluated over 3 seeds to obtain the task-averaged success rate and standard deviations.}
\centering
    \resizebox{1.\textwidth}{!}{
    \begin{tabular}{cccc cccc} 
      \toprule  
      \enspace \textbf{Method}   & Put\_block & Play\_jenga & Open\_jar &Close\_box & Open\_box & Pickup\_cup &Push\_block  \\
      \midrule
      VoxPoser~\cite{DBLP:conf/corl/HuangWZL0023} &   70.70±2.31 & 0.00±0.00  & 0.00±0.00 & 0.00±0.00 & 0.00±0.00 & 26.70±14.00  & \textbf{25.33}±8.33 
\\
      CAP~\cite{DBLP:conf/icra/LiangHXXHIFZ23} &  84.00±16.00 & 0.00±0.00 & 0.00±0.00 & 0.00±0.00 & 0.00±0.00 &  14.67±4.62 & 8.00±4.00
\\
      Scaling-up~\cite{DBLP:conf/corl/HaFS23} &  77.33±6.11  & 0.00±0.00 & 78.67±11.55 & 0.00±0.00 & 0.00±0.00 & 9.33±2.26  & 5.33±6.11
\\
      HiWE (Ours) &  \textbf{93.33}±4.16  & \textbf{82.00}±10.39 & \textbf{84.00}±6.00 & \textbf{48.67}±12.06 & \textbf{34.67}±16.77 & \textbf{88.67}±3.06 & 23.33±10.07
\\
      HiWE w/o FT &  75.33±7.57  & 60.67±9.45 & 71.33±10.07 & 31.33±9.02 & 8.67±3.06 & 74.67±6.11 & 14.67±11.02
\\
      \bottomrule 
      \toprule
        \enspace \textbf{Method}   & Take\_umbrella & Sort\_mustard & Open\_wine & Lamp\_on & Put\_knife & Pick\_\&\_lift & Insert\_block  \\
      \midrule
      VoxPoser~\cite{DBLP:conf/corl/HuangWZL0023} &   33.33±8.33 & \textbf{96.00}±6.93  & 8.00±4.00 & 57.30±12.22 &  \textbf{92.00}±4.00 & 96.00±0.00  & 0.00±0.00
\\
      CAP~\cite{DBLP:conf/icra/LiangHXXHIFZ23} &  4.00±4.00 & 0.00±0.00 & 0.00±0.00 & 64.00±6.93 & 14.67±8.33 & \textbf{ 100.00}±0.00  & 0.00±0.00
\\
      Scaling-up~\cite{DBLP:conf/corl/HaFS23} & 6.67±2.31  & 41.33±12.86 & 33.33±20.13 & 60.00±8.00 & 24.00±0.00 & \textbf{100.00}±0.00  & 0.00±0.00
\\
      HiWE (Ours) &  \textbf{64.67}±15.01  & 73.67±15.50 & \textbf{45.33}±11.02 & \textbf{72.67}±14.05 & 60.00±14.00 & 87.33±10.26  & \textbf{36.67}±5.03
\\
      HiWE w/o FT &  50.67±11.02  & 59.33±8.33 & 26.67±8.08 & 62.67±13.32 & 45.33±5.03 & 69.33±15.14 & 14.67±7.02
\\
      \bottomrule 
    \end{tabular}
    }
    \label{table:zero-shot}
\end{table*}

\begin{figure*}[t]
  \centering
  \includegraphics[width=1.\textwidth]{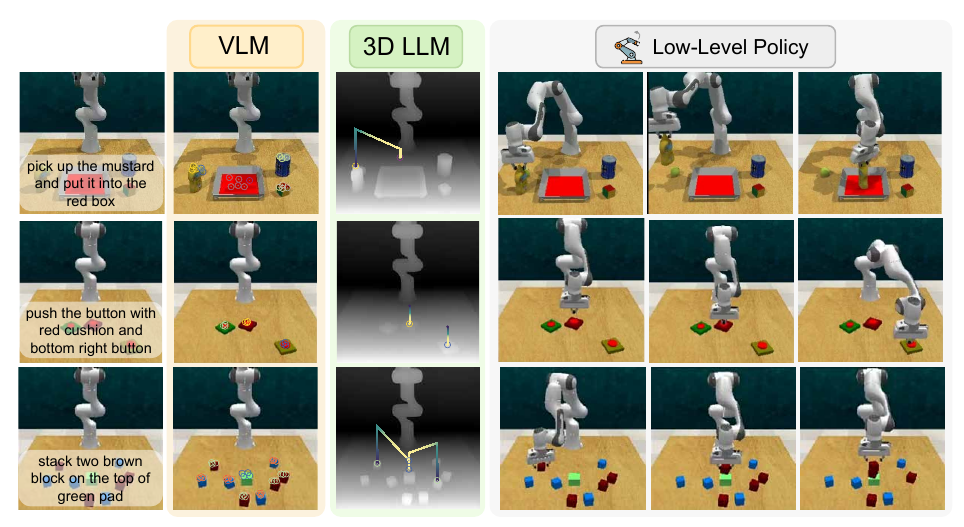}
  \caption{\textbf{Illustrative manipulation sequences.} The displayed stages connect object localization, a spatial motion plan, and robot execution for scenes involving several interactions.} 
  \label{fig:examplerollout}
\end{figure*}

\section{Experiments}

We evaluate the original HiWE configuration through task execution, point localization, component ablations, and behavior cloning from generated demonstrations. The configuration uses PointVLM and current-scene 3DLLM planning, without the ASM or KnowledgeBank extensions.

\paragraph{Relationship between the reported experiments.}
The baseline entries for VoxPoser, CAP, and Scaling-up in~\Cref{table:zero-shot}, and the CAP and RoboPoint entries in~\Cref{table:realworld}, also appear in GeneralVLA~\cite{ma2026generalvla}. They are reported here to contextualize HiWE's results and should not be interpreted as evidence of an independent replication across the two manuscripts. The tables do not provide a controlled comparison of HiWE against the later extension.

\noindent \textbf{Implementation details.} 
PointVLM extracts the object-associated coordinates used to construct the planner input.
The implementation represents each object with at least three points to expose more spatial information than a single location.
DeepSeek R1 receives the textual 3D scene description and generates the motion plan.
The output is limited to 20 spatial waypoints to bound the length of the generated plan.
Full prompts are included in the Appendix. 
PointVLM uses the front-camera view in these experiments.
The image input has a resolution of $256 \times 256$.

\begin{table}[t]
\setlength{\tabcolsep}{1.5pt}
\caption{\textbf{Quantitative comparisons on object reference (RoboRefIt). } The metric is percentage of predicted points within the target mask.}
\centering
    \small 
    \begin{tabular}{cccc c} 
      \toprule  
      \enspace \textbf{Qwen-VL}   & \textbf{LLaVA-NeXT}  & \textbf{SpaceLLaVA}   & \textbf{GPT-4o} &  \textbf{PointVLM}   \\
      \midrule
      24.1±0.9 &  20.0±0.9  & 21.3±0.9 &  15.3±1.3 &  52.1±1.2
\\

      \bottomrule 
    \end{tabular}
    \label{table:pointprediction}
\end{table}

\begin{table}[t]
\setlength{\tabcolsep}{1.5pt}
\caption{\textbf{Ablation on the data composition. }Results on RoboRefIt show that best results are achieved when all of the data sources are combined during instruction-tuning.}
\centering
    \begin{tabular}{cccc cc} 
      \toprule  
      \enspace \textbf{No VQA}   & \textbf{No LVIS}  & \textbf{No Pixel}   & \textbf{No Sim} & \textbf{No Robo}& \textbf{All}  \\
      \midrule
      42.5±3.7 &  25.8±2.1  & 32.6±5.3 &  46.6±3.2 &  47.8±3.3&  52.1±1.2
\\

      \bottomrule 
    \end{tabular}
    \label{table:datacomposition}
\end{table}

\subsection{Zero-shot performance in simulation}
\label{Zero-shot Performance in Simulation}
The simulation evaluation covers 14 tasks, including grasping and non-prehensile interactions. We measure task success under the specified environment and baseline inputs.

\noindent\textbf{Benchmark and execution.}
The 14 RLBench tasks~\cite{DBLP:journals/ral/JamesMAD20} vary in object identity, placement, and the number of required interactions. The simulator is CoppeliaSim, accessed through PyRep, with a Franka Panda arm and parallel gripper. The environment provides four RGB-D cameras; the front view supplies the PointVLM input described above. A motion planner~\cite{DBLP:journals/ram/SucanMK12} converts the requested waypoints into executable robot motion. Task-specific success conditions are listed in the appendix.

\noindent\textbf{Comparison methods and their inputs.}
We include Code-as-Policies (CAP)~\cite{DBLP:conf/icra/LiangHXXHIFZ23}, Scaling-up-Distilling-Down~\cite{DBLP:conf/corl/HaFS23}, and VoxPoser~\cite{DBLP:conf/corl/HuangWZL0023}. Their control interfaces differ: CAP composes programs from supplied action primitives; Scaling-up performs language-guided exploration with 6-DoF primitives; VoxPoser obtains motion targets from spatial value maps. In this evaluation, CAP and Scaling-up receive simulator states and object models, and VoxPoser receives segmented object point clouds. These inputs differ from HiWE's learned grounding, so the table compares the specified complete systems rather than isolating planner quality. This benchmark protocol and its baseline results are shared with the GeneralVLA report~\cite{ma2026generalvla}.

\noindent\textbf{Coverage and limitations.}
Table~\ref{table:zero-shot} reports a nonzero success rate for HiWE on every tested task. The corresponding coverage is 10 tasks for Scaling-up, 9 for VoxPoser, and 7 for CAP. HiWE has the highest reported mean on 10 tasks; the remaining tasks show that its advantage is not uniform. In particular, non-prehensile interactions and fine manipulation expose the need for more accurate pose estimates or adjustments during execution. The illustrated rollouts (\Cref{fig:examplerollout}) show examples with multiple objects and interaction stages, rather than establishing performance on every such task.

\subsection{Real-world experiments}
\noindent\textbf{Hardware and trial design.}
Physical tests use an Agilex-2.0 Piper arm with a parallel gripper and an Intel RealSense L515 LiDAR RGB-D camera viewing the workspace from above. Language instructions specify four tasks: move\_spray\_bottle, open\_drawer, open\_jar, and sort\_object. Each task has three trials of ten episodes, with object poses varied across episodes. Table~\ref{table:realworld} summarizes the resulting success rates.

\begin{table}[t]
  \centering
  \small
  \setlength{\tabcolsep}{7pt}
  \caption{\textbf{Zero-shot success rates (\%) in real-world manipulation.}
  HiWE consistently outperforms Code as Policies (CAP)~\cite{DBLP:conf/icra/LiangHXXHIFZ23} and RoboPoint across four representative tasks.}
  \label{table:realworld}
  \begin{tabular}{@{}lcccc@{}}
    \toprule
    \textbf{Method}
      & \textbf{\shortstack{Move Spray\\Bottle}}
      & \textbf{\shortstack{Open\\Drawer}}
      & \textbf{\shortstack{Open\\Jar}}
      & \textbf{\shortstack{Sort\\Object}} \\
    \midrule
    CAP       &  6.67 &  0.00 & 36.67 & 70.00 \\
    RoboPoint  &  0.00 &  0.00 & 20.00 & 63.33 \\
    HiWE       & \textbf{60.00} & \textbf{33.33} & \textbf{56.67} & \textbf{80.00} \\
    \bottomrule
  \end{tabular}
\end{table}

\noindent\textbf{Observed behavior.}
HiWE succeeds in at least some episodes of all four tasks and has a higher reported success rate than both listed baselines. The examples in~\Cref{fig:realworld} illustrate the role of spatial planning: bottle placement uses the inferred bottle height, while drawer opening requires a motion consistent with the drawer's orientation. In the evaluated baseline configurations, RoboPoint supplies localization without the trajectory-planning stage, and CAP lacks a supplied drawer-opening primitive. These implementation choices constrain what can be concluded from the comparison.

\section{Ablation Study}
\subsection{Point location accuracy of PointVLM} 

Point localization is measured by the fraction of predicted points that fall within the annotated target mask. Table~\ref{table:pointprediction} reports the mean and standard deviation across 3 runs. PointVLM obtains the highest value among the listed models; this metric evaluates localization rather than the feasibility of a complete robot motion.

Table~\ref{table:datacomposition} removes each supervision source in turn. Every removal lowers the reported RoboRefIt accuracy, with the largest decrease occurring when LVIS is omitted.
The result supports using segmentation-derived semantic point annotations in the training mixture.

\subsection{Information required by 3DLLM} 
The planner ablations vary the dimensionality of the coordinates, the number of points describing each object, and the inclusion of obstacles (\Cref{table:3DLLM}).
The 2D variant plans in image coordinates before converting the path into 3D. The one-point variant removes the geometric extent supplied by multiple points. The no-obstacle variant retains the target object but omits surrounding obstacle information. These changes test whether the available scene description is sufficient for interactions such as extracting an umbrella along the direction permitted by its stand.
The full representation performs best or ties for best on the two reported tasks. Obstacle information has a clear effect on Take umbrella, whereas its removal leaves the reported Put block result unchanged. The results therefore support task-dependent benefits rather than universal necessity of every input.



\begin{figure}[t]
  \centering
  \includegraphics[width=\linewidth]{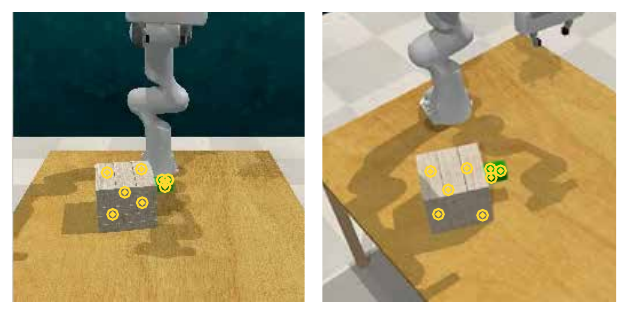}
  \caption{\textbf{The multi-view robustness of PointVLM.} This assists 3DLLM in reasoning about the direction for pulling out the block.}
  \label{fig:viewpoints}
\end{figure}

\begin{table}[htbp]
  \centering
  \caption{Ablation Study on 3DLLM for Trajectory Planning}
  \label{table:3DLLM}
  \resizebox{\columnwidth}{!}{
  \begin{tabular}{ccc}
    \toprule
    \textbf{Method}   & Take umbrella  & Put block     \\
    \midrule
    3DLLM-2D          &  2.00±2.00     & 19.33±5.03   \\
    3DLLM-1point      & 26.67±3.06     & 82.00±4.00   \\
    3DLLM w/o obstacle& 23.33±7.02     & 93.33±4.16   \\
    3DLLM             & 64.67±15.01    & 93.33±4.16   \\
    \bottomrule
  \end{tabular}
  }
\end{table}

\begin{table}[htbp]
  \centering
  \caption{Ablation Study on the HGM Module}
  \label{table:HGM}
  \resizebox{\columnwidth}{!}{
  \begin{tabular}{ccc}
    \toprule
    \textbf{Method}   & Play jenga & Take umbrella   \\
    \midrule
    HGM w/o rgb       & 56.67±7.57  & 34.00±12.49   \\
    HGM w/o 3D point  &  0.00±0.00  &  0.00±0.00    \\
    HGM w/o filter-C  & 59.33±5.03  & 54.67±15.01   \\
    HGM w/o filter-N  & 78.67±7.57  & 52.00±14.00   \\
    HGM               & 82.00±10.39 & 64.67±15.01   \\
    \bottomrule
  \end{tabular}
  }
\end{table}

\subsection{Ablation study of the HGM module} 
Table~\ref{table:HGM} tests the information and selection rules used by HGM on Play jenga and Take umbrella.
Removing RGB leaves depth-only grasp estimation. Removing the semantic 3D points eliminates the target-object guidance. The filter-C ablation disables collision rejection; the filter-N ablation samples a remaining candidate randomly instead of using the nearest-center rule.
The full HGM configuration has the highest reported mean on both tasks. Removing target-point guidance yields no successful trials in this evaluation, while the other ablations reduce success without eliminating it. These findings concern the tested tasks and do not imply that each modality is indispensable for every manipulation problem.

\subsection{Data scaling}
We examine how the performance of an RVT-2 policy changes as more HiWE demonstrations are supplied for training. The comparison uses demonstrations generated by RLBench as a reference.
The reported linear fits have slopes of 0.543 for HiWE-generated demonstrations and 0.156 for RLBench-generated demonstrations. These coefficients summarize the evaluated range and are not a general scaling law.




\section{Conclusion}
HiWE establishes a point-based pipeline for connecting visual grounding, spatial language planning, and grasp execution. The original system uses a fine-tuned PointVLM to produce geometric anchors, 3DLLM to plan from the current scene, and HGM to resolve grasp poses. Its evaluation examines manipulation success, the visual supervision mixture, and the spatial information needed for planning. Successful executions also provide demonstrations for policy training. GeneralVLA subsequently extends this framework through affordance segmentation and persistent experience retrieval~\cite{ma2026generalvla}; these mechanisms are outside the HiWE configuration evaluated here. Remaining limitations include errors in point localization, incomplete geometric descriptions, and grasp execution failures. Zero-shot deployment should be understood in the stated task-specific sense, since existing robot observations are part of the visual training data.

\bibliography{main}

\label{main:end}


\end{document}